\documentclass[journal]{IEEEtran}

\usepackage{cite}
\usepackage{amsmath,amssymb,amsfonts}
\usepackage{algorithmic}
\usepackage{graphicx}
\usepackage{textcomp}
\usepackage{hyperref}
\usepackage{wrapfig}
\usepackage{xcolor}
\usepackage{soul}
\usepackage{gensymb}

\begin{document}

\newcommand{\datasetUrl}{\url{https://github.com/ansfl/MoRPI}} 
\newcommand{\hlc}[2][white]{{%
    \colorlet{foo}{#1}%
    \sethlcolor{foo}\hl{#2}}%
}

\title{MoRPI: Mobile Robot Pure Inertial Navigation}
\author{Aviad~Etzion, and~Itzik~Klein,~\IEEEmembership{Senior Member,~IEEE}
\thanks{A. Etzion and I. Klein are with the Hatter Department of Marine Technologies, University of Haifa, Israel.}
\thanks{Manuscript received May 2022}}

\maketitle

\begin{abstract}
Mobile robots are used in industrial, leisure, and military applications.  In some situations, a robot navigation solution relies only on inertial sensors and as a consequence, the navigation solution drifts in time.
In this paper,  we propose the MoRPI framework, a mobile robot pure inertial approach. Instead of travelling in a straight line trajectory, the robot moves in a periodic motion trajectory to enable peak-to-peak estimation. In this manner, instead of performing three integrations to calculate the robot position in a classical inertial solution, an empirical formula is used to estimate the travelled distance. Two types of MoRPI approaches are suggested, where one is based on both accelerometer and gyroscope readings while the other is only on gyroscopes. Closed form analytical solutions are derived to show that MoRPI produces lower position error compared to the classical pure inertial solution. In addition, to evaluate the proposed approach, field experiments were made with a mobile robot equipped with two types of inertial sensors. In total, \hlc{143 trajectories with a time duration of 75 minutes} were collected and evaluated. 
The results show the benefits of using our approach. To facilitate further development of the proposed approach, both dataset and code are publicly available at  \datasetUrl. 
\end{abstract}


\IEEEpeerreviewmaketitle

\section{Introduction}
\IEEEPARstart{M}{obile} robots are used in different applications while operating under various constraints. For example, they can be found in industry, hotels, and warehouses and can  be used for delivery, agriculture, healthcare, and military applications. Besides the improvements in technology, price reductions for electronic sensors and devices have caused an increase in research and in demand. Therefore, many companies worldwide produce mobile robots to answer the demand and infiltrate new markets.
\\
In parallel, major breakthroughs in low cost inertial sensors based on micro-electrical-mechanical-system (MEMS) technology provide better accuracy and robustness.  The inertial sensors—namely, the accelerometers and gyroscopes—are packed in an inertial measurement unit (IMU) that is relatively very small (mm in scale), has low power consumption, and can be deployed easily in a variety of devices. In pure inertial navigation, the inertial measurements are integrated to obtain the position, velocity, and orientation of the platform. However, as the inertial sensor measurements contain noise and other types of errors when integrated,  they cause the navigation solution to drift over time.\\
To compensate for such drift, external sensors or vehicle constraints were suggested in the literature, and solutions have been proposed over the years. One method, described by \cite{Borenstein1997}, is the use of absolute position measurements to obtain the location. A common approach is to use vision for navigation, \cite{vision2002}, \cite{chatterjee2013mobile}. Similarly, Lidar \cite{lidar2018}, and Sonar \cite{Leonard2012} can be used for localization. Using these methods, a pre-stored map can be saved in the robot's memory, and the robot compares its location to the saved map. Another common way to use the sensors is by scanning the environment and creating a map so the robot can estimate its relative position to features and seek landmarks (SLAM) \cite{SLAM2004}. A disadvantage of these methods is the sensitivity to changes. For example, when furniture is moved, this can confuse the navigation system. Moreover, reflections or a lack of proper light can blind the sensors. \\
Another approach is to use active beacons. Antennas, placed in known locations, can cover the environment so that triangulation \cite{Triangulation2003} or trilateration \cite{Trilateration2005} can be used to compute locations. GNSS is an example of this kind of navigation sensor. In order to obtain the location with beacons, they need to be situated in known locations, and eye contact between the robot receiver and the beacons is mandatory. Therefore, GPS cannot be used indoors, in urban canyons, or in outer space. 
\\
Another kind of method is dead-reckoning. The IMU belongs to this group together with odometry. In odometry, sensors are installed next to the wheels \cite{Odometry1996,Odometer2021}. Relying on the pre-determined wheel diameter and wheelbase, the position and heading are achieved. 
\\
In some situations, external measurements are not available and the solution is based only on inertial sensors; hence, the navigation solution drifts in time. For example, GNSS signals are not available indoors and cameras suffer from lighting conditions. To cope with such situations, vehicle constraints could be applied. Several approaches were presented over the years using different types of prior knowledge as pseudo-measurements. For example, model of the vehicle dynamics and operating environment such that the vehicle travelling on a road \cite{klein2010pseudo,brandt1998constrained}, using stationary updates for zero velocity and angular velocity \cite{ramanandan2011inertial}, and modelling the sensor error \cite{Barshan1995}.
\\
In other navigation domains such as indoors, to cope with the navigation solution drift, instead of integrating  the inertial sensor readings, an empirical formula estimates the drift in a pedestrian dead reckoning (PDR) framework \cite{PDR2011}. In recent years, such empirical formulas have been replaced by machine learning approaches to regress the change in distance in any required time interval \cite{pdrnet2021,yan2018ridi}. 
Recently, the quadrotor dead reckoning (QDR) framework was developed for pure inertial navigation of quadrotors, employing PDR guidelines to improve position accuracy \cite{Shurin2020}.\\  
In this paper, inspired by PDR and QDR, we derive the MoRPI framework: a mobile robot pure inertial navigation solution that operates for short time periods to bound the navigation solution drift when external sensors are not available. The main idea is to drive the robot in a periodic motion instead of a straight line trajectory, \hlc{as commonly the path planning of mobile robots made up of straight lines,} and adjust some of the PDR and QDR principles. This is done in MoRPI-A where both accelerometer and gyroscopes readings are used to determine the robot's two-dimensional position. In some scenarios\hlc{, like narrow corridors when the amplitude should be small,} the periodic motion may not be reflected in the accelerometer readings due to their high noise characteristics, so we also offer MoRPI-G, which uses only the gyroscope measurements to calculate the position of the robot. The contributions of this paper:
\begin{enumerate}
    \item The MorRPI framework copes with situations of pure inertial navigation in mobile robots.
    \item MoRPI-G determines the mobile robot position using only gyroscope measurements.
    \item An analytical error assessment of the MoRPI approach is provided and compared to the classical pure inertial solution.
    \item Our dataset and code are publicly available and can be found here: \datasetUrl. 
\end{enumerate}
\par
To evaluate the proposed approach, field experiments were made with a mobile robot equipped with two types of inertial sensors. In total, \hlc{143 trajectories with a time duration of 75 minutes} were collected and evaluated. Comparisons to the classical inertial navigation solution were made in two and three dimensions. \\
The rest of the paper is organized as follows: Section \ref{sec:PF} presents the INS equations and the QDR method. Section \ref{sec:PA} describes the proposed MoRPI approach and provides an  analytical assessment of its position error. Section \ref{sec:AR} explains the experiments and gives the results, and Section \ref{sec:con} gives the conclusions of this paper.    
\section{Problem Formulation}\label{sec:PF}
In this section, we address the process of inertial measurements within the inertial navigation system (INS) to calculate the robot navigation solution in three dimensions. Also, as mobile robots move in two dimensions, the INS equations are reduced to planar motion and presented here. Then, we briefly review the QDR approach.
\subsection{Inertial Navigation System}
The INS equations provide a solution for the position, velocity, and attitude  based on the inertial sensor readings. As short time scenarios are addressed, the inertial frame (i-frame) is defined at the robot's starting point, and the body frame (b-frame) coincides with the inertial sensors' sensitive axes. Let the accelerometer measurement vector, the specific force vector expressed in the body frame $\boldsymbol{f}_{ib}^b$, be denoted as
\begin{equation}
\boldsymbol{f}_{ib}^b=\begin{bmatrix}
            f_x\\
            f_y\\
            f_z
\end{bmatrix}
\end{equation}
and the gyroscope measurement vector, the angular velocity vector expressed in the body frame $\boldsymbol{\omega}_{ib}^b$, as
\begin{equation} \label{eq:gyro_meas}
\boldsymbol{\omega}_{ib}^b=\begin{bmatrix}
            \omega_x\\
            \omega_y\\
            \omega_z
\end{bmatrix}
\end{equation}
where the subscript $ib$ stands for the body frame with respect to the inertial frame, and the superscript $b$ denotes that the vector is resolved along the axes of the body frame.\\
As our scenarios include low-cost inertial sensors and short time periods, the earth turn rate and the transport rate are neglected. Hence, the INS equations of motion are \cite{Titterton2004}: 
\begin{align}
            \dot{\boldsymbol{p}}^n & =\boldsymbol{v}^n\label{ins_eq1}\\
            \dot{\boldsymbol{v}}^n & =\mathbf{C}_b^n\boldsymbol{f}_{ib}^b+\boldsymbol{g}^n\label{ins_eq2}\\
            \dot{\mathbf{C}}_b^n & =\mathbf{C}_b^n\mathbf{\Omega}_{ib}^b\label{ins_eq3}
\end{align}
where $\boldsymbol{p}^n$ is the position vector expressed in the navigation frame, $\boldsymbol{v}^n$ is the velocity vector expressed in the navigation frame, $\boldsymbol{g}^n$ is the gravity vector expressed in the navigation frame and assumed constant throughout the trajectory, $\mathbf{C}_b^n$ is the body to navigation orthonormal transformation matrix, \hlc{and  $\mathbf{\Omega}_{ib}^b$ is the skew-symmetric matrix of the angular rate, defined as:}
\begin{equation}
    \mathbf{\Omega}_{ib}^b=\begin{bmatrix}
                            0 & -\omega_z & \omega_y\\
                            \omega_z & 0 & -\omega_x\\
                            -\omega_y & \omega_x & 0
                            \end{bmatrix}
\end{equation}
\hlc{where $\omega_{j=x,y,z}$ are the gyroscope measurements as defined in equation} \eqref{eq:gyro_meas}.
\subsection{Two-Dimensions INS}\label{sec:2dins}
Leveraging the wheeled robot planner motion, it is assumed the robot moves with nearly zero roll and pitch angles and only the motion in the $x-y$ plane is relevant. Therefore, the body-to-navigation transformation matrix depends only on the yaw angle, $\psi$, and is given by \cite{Groves2015}: 
\begin{equation}
    \mathbf{C}_b^n=\begin{bmatrix}\label{eq:tbnr}
            \cos\psi & -\sin\psi & 0\\
            \sin\psi & \cos\psi & 0\\
            0 & 0 & 1
    \end{bmatrix}.
\end{equation}
Substituting \eqref{eq:tbnr} into \eqref{ins_eq2} shows that $f_z$ has no influence on the velocity and the position in the $x-y$ plane, and thus it is not needed in the inertial calculation.  In addition, as only the yaw angle is taken into account, the gyro measurements in the $x-y$ plane, i.e., $\omega_x, \omega_y$, are neglected and only $\omega_z$ is considered. 
\subsection{Quadrotor Dead Reckoning}
In \cite{Shurin2020}, an adaptation of PDR principles was used to derive the QDR approach for situations of pure inertial navigation for quadrotors. To that end,  the accelerometer readings were used to detect a peak-to-peak event. Then, using a step length estimation approach, the peak-to-peak distance was estimated. In their analysis, the Weinberg approach \cite{Weinberg2002} was employed to estimate the peak-to-peak distance. \hlc{Originally, it was developed to cope with constant stride length estimation approaches (based on user height). To that end, Weinberg proposed an empirical method taking into account the accelerometer readings during each stride. The underlying assumption of this approach is that the vertical bounce (impact) is proportional to the stride length. In the QDR approach, the peak-to-peak distance estimation is:}
\begin{equation}
            s_w=G_w\big(\max\left(f_{ib}^b\right)-\min\left(f_{ib}^b\right)\big)^\frac{1}{4} \label{weinberg}
\end{equation}
where $s_w$ is the estimated peak-to-peak distance according to Weinberg's approach, and $G_w$ is the approach's gain.\\ 
To apply \eqref{weinberg}, the approach's gain needs to be determined prior to application.  Once the peak-to-peak distance is found, it is used together with the gyro-based heading and initial conditions to propagate the quadrotor position by
\begin{align}
            x_{k+1}=x_k+s_k\cos\Delta\psi_k \label{x_dead_reckon}\\
            y_{k+1}=y_k+s_k\sin\Delta\psi_k \label{y_dead_reckon}
\end{align}
where $k$ is the time index.
\section{Proposed Approach}\label{sec:PA}
\begin{figure}[!htbp]
        \centering
        \includegraphics[width=3in]{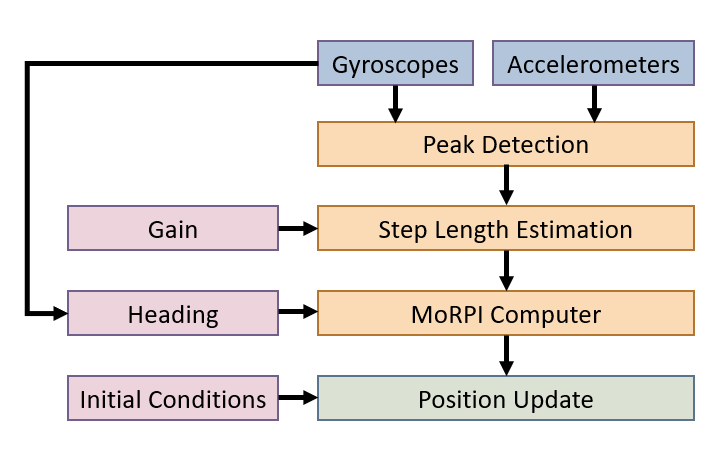}
        \caption{MoRPI framework for pure inertial navigation of mobile robots.}
        \label{fig_scheme}
\end{figure}

Motivated by the QDR approach, our goal is to derive an accurate navigation solution for mobile robots using only inertial sensors for short time periods. Compared to the QDR approach, the mobile robot maneuvers are limited due to the indoor environment (corridors, for example). As a consequence, the periodic motion requires fewer accelerations, which may not be sensed using low-cost MEMS accelerometers. To cope with this challenge, in addition to applying and modifying QDR for mobile robots (MoRPI-A), we propose a gyroscope-only solution for positioning the mobile robot (MoRPI-G). We argue that regardless of the limited space for maneuvering, the angular rate in the $z$ direction (perpendicular to the robot's plane of motion) is dominant enough to be recognized and utilized for positioning the robot. \\
Both of Our MoRPI approaches consist of the following phases:
\begin{itemize} 
    \item \textbf{Peak detection}: The peaks during the motion are extracted as local maxima from the inertial measurements.
    \item \textbf{Gain calculation}: Prior to the application of the proposed approach, the empirical gain is estimated by moving the robot at a known distance with a known number of periods while using the Weinberg approach. This procedure is repeated several times with slightly different maneuvers and the gain is taken as the average from all runs. Once obtained, this gain is used in real-time to estimate the peak-to-peak distance.
    \item \textbf{Peak-to-peak distance estimation}: The 'step', in analogue to PDR, is the segment between two peaks. The peak-to-peak distance estimation is done using the Weinberg approach with the predefined gain and the inertial sensor readings.
    \item \textbf{Heading determination}: We use the heading extracted from the transfer matrix $\mathbf{C}_b^n$  to project the peak-to-peak distance into local planar coordinates.
    \item \textbf{Position update}: As a dead-reckoning method, the position is updated relative to the previous step while using the current heading angle and peak-to-peak distance.
\end{itemize}
Our proposed approach is illustrated in Figure~\ref{fig_scheme}.\\
As discussed above, we distinguish between two MoRPI approaches based on the inertial sensors they employ for the peak-to-peak length estimation: 
\begin{enumerate}
    \item \textbf{MoRPI-A}: uses both accelerometers and gyroscopes. As applied in PDR and QDR,  the advantage of this method over the INS is that it uses less integration on the inertial sensor readings and as a result reduces the position drift. \hlc{For clarity, we define the body coordinate frame axes: the x-axis points towards the moving direction, the z-axis points downwards, and the y-axis completes the orthogonal set.} In PDR, the motion is expressed in the vertical direction; thus, the accelerometer $z$-axis readings are used to determine the step length. In QDR, the magnitude of the specific force vector is used instead. In the proposed approach, the $y$-axis accelerometer readings are used instead, as the applied periodic motion is exhibited and captured  best in this direction. Thus, the peak-to-peak distance is calculated by
    \begin{equation}
            s_A=G_A\big(\max\left(f_y\right)-\min\left(f_y\right)\big)^\frac{1}{4} \label{eq:morpi-a}
    \end{equation}
    where $s_A$ is the peak-to-peak distance and $G_A$ is the gain of MoRPI-A. \hlc{In general, it is necessary to determine $G_A$ before using} \eqref{eq:morpi-a}. \hlc{To that end, the mobile robot is moved in a trajectory with the required dynamics, where the travelled distance of this trajectory is known.  By plugging the accelerometer readings in each peak-to-peak distance and summing the results, the gain value can be estimated. Commonly, this procedure is repeated to obtain a more accurate gain.}
    \item \textbf{MoRPI-G}: uses only  gyroscopes. To cope with real-world situations of small amplitudes within the periodic motion (in the horizontal plane) that cannot be sensed by the accelerometers, we employ the gyro z-axis readings for estimating the robot's peak-to-peak distance using
    \begin{equation}
            s_G=G_G\big(\max\left(\omega_z\right)-\min\left(\omega_z\right)\big)^\frac{1}{4}
            \label{eq:morpi-g}
    \end{equation}
    where $s_G$ is the peak-to-peak distance and $G_G$ is the gain of MoRPI-G. \hlc{$G_G$ is extracted in the same manner as $G_A$ except here, instead of the accelerometer readings, the angular rate $\omega_z$ is employed.}
\end{enumerate}
Regardless of how the peak-to-peak distance was estimated, i.e., by \eqref{eq:morpi-a} or \eqref{eq:morpi-g}, the robot position is calculated by
\begin{align}
            x_{k+1}=x_k+s_{i,k}\cos\Delta\psi_k \label{eq:morpi_x}\\
            y_{k+1}=y_k+s_{i,k}\sin\Delta\psi_k \label{eq:morpi_y}
\end{align}
where $i={A,G}$ depending on the approach. As relative positioning is used here, the initial position is set to zero.
\subsection{Analytical Assessment} \label{subsec:AA}
\begin{figure}[!htbp]
            \centering
            \includegraphics[width=3.5in]{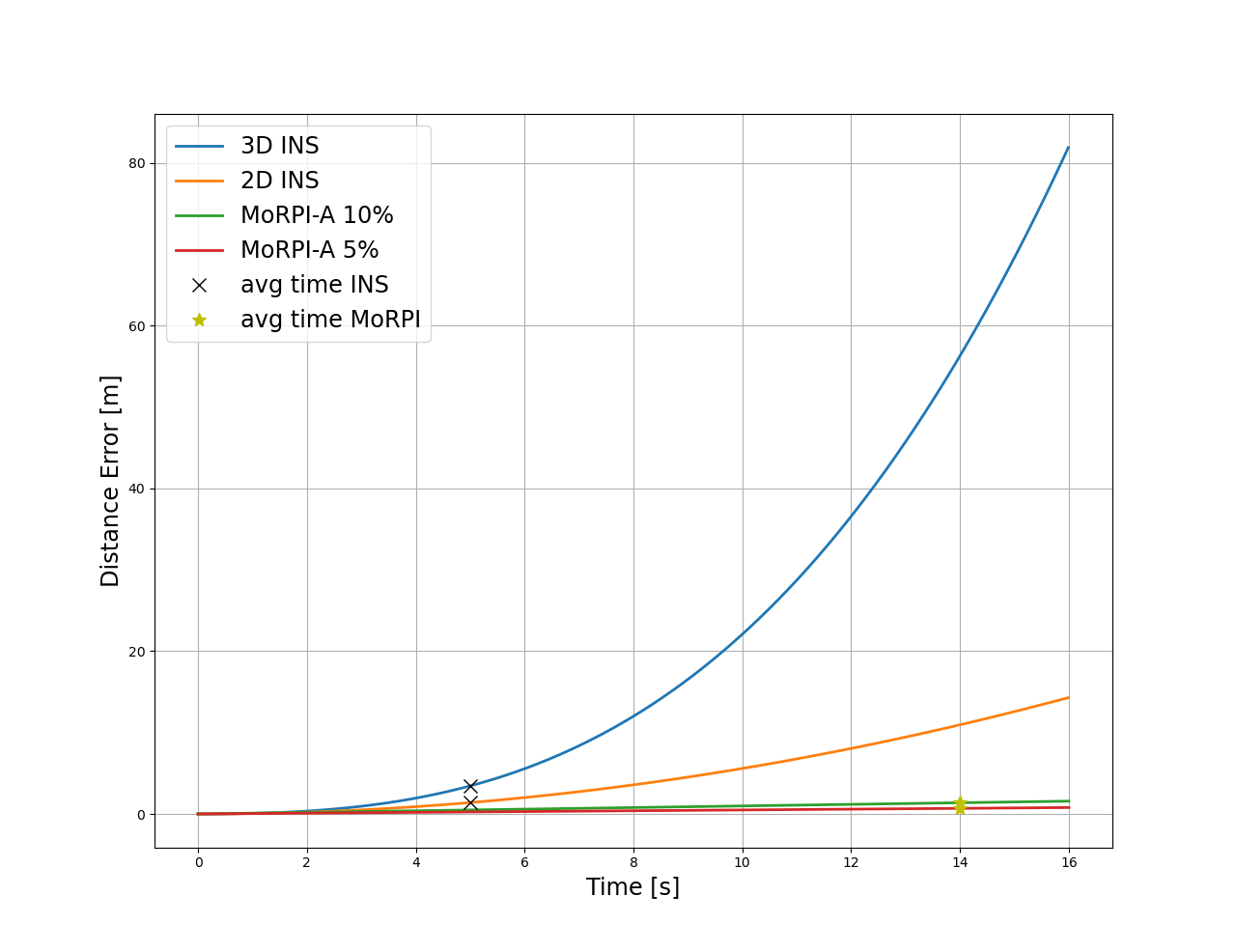}
            \caption{Analytical assessment of 3D \& 2D INS vs. MoRPI-A distance error.}
            \label{fig_assessment}
        \end{figure}

In this section, we offer an analytical assessment of the expected position error using two and three dimensional INS while the robot moves in a straight line trajectory compared to our proposed approach where the robot moves in a periodic motion trajectory with the same distance. Maintaining consistency with Section~\ref{sec:PF}, the earth and transport rates are neglected in the analysis. \\
We employ the 15 error state model  \cite{farrell2008,groves2015principles} expressed in the navigation frame  with the following error state vector:
        \begin{equation}
            {\delta}\boldsymbol{x}=\begin{bmatrix}{\delta}\boldsymbol{p}^n & {\delta}\boldsymbol{v}^n & \boldsymbol{\epsilon}^n & \boldsymbol{b}_a & \boldsymbol{b}_g \end{bmatrix}^T
        \end{equation}
where ${\delta}\boldsymbol{p}^n$ is the position error vector expressed in the navigation frame, ${\delta}\boldsymbol{v}^n$ is the velocity error vector expressed in the navigation frame, $\boldsymbol{\epsilon}^n$ is the  misalignment vector, $\boldsymbol{b}_a$ is the accelerometer bias residuals, and the gyro bias residuals is $\boldsymbol{b}_g$, as expressed in the body frame. As short time periods are considered, we assume constant biases during the analysis. \\
The resulting error state model is 
        \begin{equation} \label{eq:lin_sys}
           {\delta}\dot{\boldsymbol{x}}=\mathbf{F}{\delta}\boldsymbol{x}
        \end{equation}
where $\mathbf{F}$ is the system matrix
        \begin{equation}
            \mathbf{F}=\begin{bmatrix}
                        \mathbf{0}_{3\times3} & \mathbf{I}_{3} & \mathbf{0}_{3\times3} & \mathbf{0}_{3\times3} & \mathbf{0}_{3\times3} \\
                        \mathbf{0}_{3\times3} & \mathbf{0}_{3\times3} & -\left(\boldsymbol{f}^n\times\right) & \mathbf{C}_b^n & \mathbf{0}_{3\times3}\\
                        \mathbf{0}_{3\times3} & \mathbf{0}_{3\times3} & \mathbf{0}_{3\times3} & \mathbf{0}_{3\times3} & \mathbf{C}_b^n\\
                        \mathbf{0}_{3\times3} & \mathbf{0}_{3\times3} & \mathbf{0}_{3\times3} & \mathbf{0}_{3\times3} & \mathbf{0}_{3\times3}\\
                        \mathbf{0}_{3\times3} & \mathbf{0}_{3\times3} & \mathbf{0}_{3\times3} & \mathbf{0}_{3\times3} & \mathbf{0}_{3\times3}
                        \end{bmatrix},
        \end{equation}
and $-\left(\boldsymbol{f}^n\times\right)$ is the skew-symmetric form of the specific force vector expressed in the navigation frame.\\
The solution of the set of first order differential equations \eqref{eq:lin_sys} is
        \begin{equation}\label{eq:state_sol}
            {\delta}\boldsymbol{x}(t)=\mathbf{\Phi}{\delta}\boldsymbol{x}_{t_0}
        \end{equation}
where ${\delta}\boldsymbol{x}_{t_0}$ is the initial condition vector of the system and $\mathbf{\Phi}$ is the transition matrix.
A closed form solution of the transition matrix in \eqref{eq:state_sol} was offered in \cite{ramanandan2011observability,Klein2020}: 
 \begin{equation} \label{eq:phi}
            \mathbf{\Phi}(t,t_0)=\begin{bmatrix}
                                    \mathbf{I}_3 & \left(t-t_0\right)\mathbf{I}_3 & \mathbf{P}_t & \mathbf{Q}_t & \mathbf{T}_t\\
                                    \mathbf{0}_{3\times3} & \mathbf{I}_3 & \mathbf{S}_t & \mathbf{R}_t & \mathbf{M}_t\\
                                    \mathbf{0}_{3\times3} & \mathbf{0}_{3\times3} & \mathbf{I}_3 & \mathbf{0}_{3\times3} & \mathbf{R}_t\\
                                    \mathbf{0}_{3\times3} & \mathbf{0}_{3\times3} & \mathbf{0}_{3\times3} & \mathbf{I}_3 & \mathbf{0}_{3\times3}\\
                                    \mathbf{0}_{3\times3} & \mathbf{0}_{3\times3} & \mathbf{0}_{3\times3} & \mathbf{0}_{3\times3} & \mathbf{I}_3
                                 \end{bmatrix}
        \end{equation}
where 
        \begin{align}
            \mathbf{P}_t&=\int_{t_0}^{t} \mathbf{S}_s \,ds \quad  &\mathbf{S}_t&=-\int_{t_0}^{t} \left(\boldsymbol{f}^n\times\right) \,ds\\
            \mathbf{Q}_t&=\int_{t_0}^{t}\mathbf{R}_s \,ds \quad &\mathbf{R}_t&=\int_{t_0}^{t} \mathbf{C}_b^n(\tau) \,d\tau\\
            \mathbf{T}_t&=\int_{t_0}^{t} \mathbf{M}_s \,ds \quad &\mathbf{M}_t&=-\int_{t_0}^{t} \left(\boldsymbol{f}^n\times\right)\mathbf{R}_s \,ds
        \end{align}
As a straight line trajectory for short time periods is considered, we assume that the body and navigation frame coincide:
\begin{equation} \label{eq:asp1}
   \mathbf{C}_b^n=\mathbf{I}_3,    
\end{equation}
and, as a consequence 
\begin{equation} \label{eq:asp2}
   \boldsymbol{f}^n\times=\begin{bmatrix}
                    0 & -\left(g+b_{a,z}\right) & b_{a,y}\\
                    g+b_{a,z} & 0 & -b_{a,x}\\
                    -b_{a,y} & b_{a,x} & 0
                  \end{bmatrix}    
\end{equation}
where $b_{a,x}, b_{a,y}$, and $b_{a,z}$ are the biases of the accelerometer in the $x$, $y$, and $z$ axes, respectively. Finally, as for both INS and MoRPI approaches the initial position and misalignment errors have the same influence on the position error, we assume zero initial position and misalignment errors. Yet, the initial velocity error influences only  the INS approaches due to the integration on the velocity states. In the MoRPI approach, the position is obtained from an empirical formula without the need to integrate velocity errors. Thus, only the initial velocity error
\begin{equation} \label{eq:asp3}
  {\delta}\boldsymbol{v}(t=0)= {\delta}\boldsymbol{v}_{t_0}
\end{equation}
is considered in our analysis. \\
Taking into account \eqref{eq:asp1}-\eqref{eq:asp3}, when solving \eqref{eq:phi}, the position error is:
\begin{align}
    {\delta}p_x&={\delta}v_{t_0,x}t+\frac{1}{2}b_{a,x}t^2-\frac{1}{6}(g+b_{a,z})b_{g,y}t^3+\frac{1}{6}b_{a,y}b_{g,z}t^3\\
    {\delta}p_y&={\delta}v_{t_0,y}t+\frac{1}{2}b_{a,y}t^2+\frac{1}{6}(g+b_{a,z})b_{g,x}t^3+\frac{1}{6}b_{a,x}b_{g,z}t^3
\end{align}
The heading error is the same for the all methods we examined. Therefore, the elements that depend on $b_{g,z}$ were discarded. The resulting distance error is:
\begin{equation} \label{eq:error-3d}
\begin{split}
    e_{3D}=&\bigg\{\left({\delta}v_{t_0,x}^2+{\delta}v_{t_0,y}^2\right)t^2+\left({\delta}v_{t_0,x}b_{a,x}+{\delta}v_{t_0,y}b_{a,y}\right)t^3\\
    &+\big[\frac{1}{4}\left(b_{a,x}^2+b_{a,y}^2\right)-\frac{1}{3}\alpha\left({\delta}v_{t_0,x}b_{g,y}+{\delta}v_{t_0,y}b_{g,x}\right)\big]t^4\\
    &-\frac{1}{6}\alpha\left(b_{a,x}b_{g,y}+b_{a,y}b_{g,x}\right)t^5\\
    &+\frac{1}{36}\alpha^2\left(b_{g,y}^2+b_{g,x}^2\right)t^6\bigg\}^{\frac{1}{2}}.
\end{split}
\end{equation}
where $\alpha \triangleq g+b_{a,z}$.\\
When considering the two dimensional INS $b_{g,x},b_{g,y}$ and $b_{a,z}$ are not relevant for the position estimation, thus the distance error \eqref{eq:error-3d} reduces to
\begin{equation} \label{eq:error-2d}
\begin{split}
    e_{2D}=&\bigg\{\left({\delta}v_{t_0,x}^2+{\delta}v_{t_0,y}^2\right)t^2+\left({\delta}v_{t_0,x}b_{a,x}+{\delta}v_{t_0,y}b_{a,y}\right)t^3\\&+\frac{1}{4}\left(b_{a,x}^2+b_{a,y}^2\right)t^4\bigg\}^{\frac{1}{2}}.
\end{split}
\end{equation}
As a consequence, the expected error of the two-dimensional INS is smaller than the three dimensional one. \\
In our MoRPI approaches, the distance error is based on the peak-to-peak distance based on the Weinberg approach \eqref{eq:morpi-a} 
for MoRPI-A and \eqref{eq:morpi-g} 
for MoRPI-G. In this analysis, we focus only on MoRPI-A as the same procedure  can be applied exactly on MoRPI-G. 
As shown in  \eqref{eq:morpi-a},
the peak-to-peak distance is a function of the gain and the specific force readings in the y-axis between the peaks. Let 
\begin{equation}\label{eq:del_F}
    \Delta f=\big(\max\left(f_y\right)-\min\left(f_y\right)\big)^\frac{1}{4}.
\end{equation}
Note that as constant biases are addressed they are cancelled out in \eqref{eq:del_F} and therefore have no influence on the distance error. \\
Substituting \eqref{eq:del_F} into \eqref{eq:morpi-a} 
and linearizing to obtain the peak-to-peak error at peak $k$ gives 
\begin{equation}\label{eq:after_lin}
         s_{A,k}+{\delta}s_k=G_A\Delta f+{\delta}G_A\Delta f
\end{equation}
where $s_{A,k}$ is the true $k$-th peak-to-peak value, ${\delta}s_k$ is the peak-to-peak error, and ${\delta}G_A$ reflects the error of the actual gain that should have been applied, depending on the actual periodic motion, which differs from the expected one.\\
Removing the true values of \eqref{eq:morpi-a} 
in \eqref{eq:after_lin} yields
\begin{equation}\label{eq:after_lin}
         {\delta}s_k = {\delta}G_A\Delta f.
\end{equation}
That is, the peak-to-peak distance error of the MoRPI-A approach depends only on the gain error and not on the biases of the accelerometers. The distance error of the whole trajectory is the sum of all peak-to-peak distance errors: 
\begin{equation}
            {\delta}s=\sum_{k=1}^{N} {\delta}G_A\Delta f
\end{equation}
where N is the number of peaks.\\
To summarize, the 3D INS distance error \eqref{eq:error-3d} 
and the 2D INS distance error \eqref{eq:error-2d} 
are polynomial in time and therefore expected to diverge much faster than the MoRPI-A  approach. This is illustrated in Figure \ref{fig_assessment} using numerical values as described later in Section \ref{sec:AR}. \hlc{In addition, the gain choice should  fit the expected dynamics to obtain the best performance, otherwise performance degradation should be expected} \cite{klein2018gain}.  \hlc{Hence, in practice moving the vehicle  differently than planned, will yield a position error. This behaviour corresponds to working with erroneous gain instead of the expected one. Therefore,}
to evaluate the gain error, we used ${\delta}G_A=5\%$ and ${\delta}G_A=10\%$ from the true gain. 
If the time duration of the trajectory is equal in both 2D and 3D INS (straight line) and MoRPI approaches (periodic motion) the improvement at the end of the trajectory is $2.95$m and $0.91$m with ${\delta}G_A=10\%$ for 3D and 2D, respectively. MoRPI-A with ${\delta}G_A=5\%$ improves 3D INS by $3.20$m and 2D INS by $1.15$m for $5$s trajectories.
\\
When considering the same distance and speed, in general, the straight line trajectory is faster than the periodic trajectory. Based on our dataset (Section \ref{sec:AR}) we observe that the average time of the straight line trajectory was approximate $5$s while with the MoRPI approach $14$s.
Yet, still, the MoRPI approach obtained the best performance. MoRPI-A with ${\delta}G_A=10\%$  improves 3D INS by $2.07$m and 2D INS by $0.02$m at the end of the trajectory while the improvement of MoRPI-A with ${\delta}G_A=5\%$ is $2.76$m and $0.71$m for 3D INS and 2D INS, respectively. \\
\section{Analysis and Results}\label{sec:AR}
\subsection{Field Experiment Setup}
\begin{figure}[!h]
            \centering
            \includegraphics[width=2.5in]{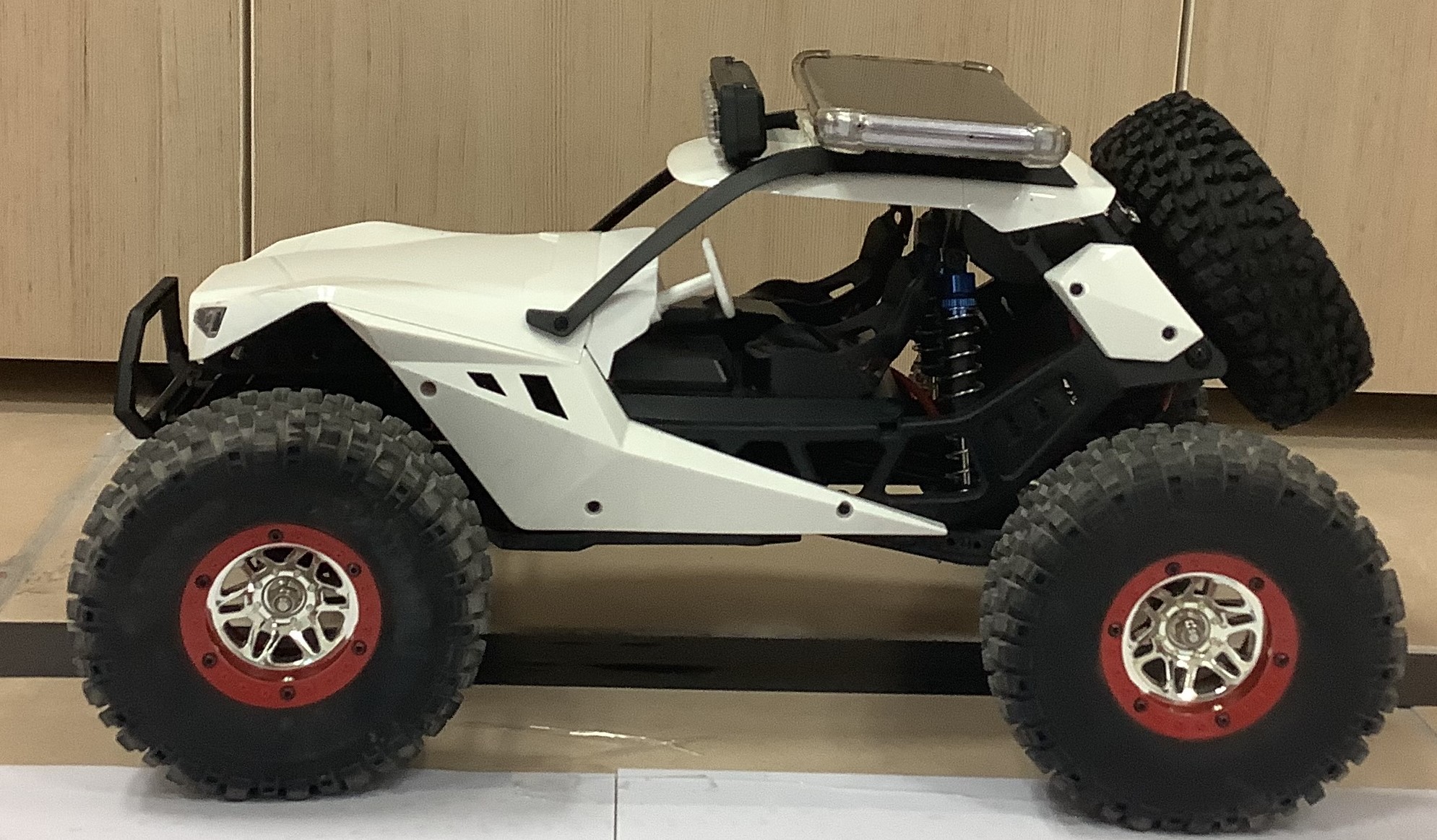}
            \caption{Setup of the RC car and the phone.}
            \label{fig_car}
        \end{figure}

A remote control car and a smartphone were used to perform the experiments and record the inertial data to create our dataset. The smartphone was rigidly attached to the car as shown in Figure~\ref{fig_car}. The model of the RC car we used is a STORM Electric 4WD Climbing car. The car dimensions are $385\times260\times205 mm$ with a wheelbase of $253 mm$ and tire diameter of $110 mm$. The car has a realistic suspension system that enables it to reach up to $40 kph$ and cross rough terrain. \\
Two different smartphones, with different inertial sensors, were used in our experiments: 
\begin{enumerate}
    \item A Samsung Galaxy S8 Smartphone with an IMU model of LSM6DSL manufactured by STMicroelectronics. 
    \item A Samsung Galaxy S6 smartphone with an IMU model of MPU-6500 manufactured by TDK InvenSense.
\end{enumerate}
\hlc{The error parameters of both sensors are presented in Table} \ref{tab:sensors param}. In both smartphones, the inertial sensor readings were recorded with a sampling rate of $100$Hz.
\begin{table}[!htb]
\renewcommand{\arraystretch}{1.3}
    \caption{Sensors Errors according to manufacturer.}
    \label{tab:sensors param}
    \centering
    \begin{tabular}{|c | c | c | c | c |} 
         \hline
            & \multicolumn{2}{| c |}{Gyro} & \multicolumn{2}{| c |}{Accelerometer} \\
        \hline
            Sensor & Bias & Noise & Bias & Noise\\ 
        \hline
         MPU-6500 & $6 \degree/s$ & $0.01 \degree/s/\sqrt{Hz}$ & $60 mg$ & $300 \micro{g}/\sqrt{Hz}$\\ 
        \hline
         LSM6DSL & $3 \degree/s$ & $0.004 \degree/s/\sqrt{Hz}$ & $40 mg$ & $130\micro{g}/\sqrt{Hz}$\\ 
        \hline
    \end{tabular}
\end{table}
\\
The smartphone was placed on the top of the car with the screen facing upward. 
At the starting point, the car was directed to the end point and the phone accelerometer on the $x$-axis was aligned to the direction of movement. At the beginning of each recording, the phone was mounted parallel to the floor.
\subsection{Dataset}
\hlc{Five} types of trajectories were made during the field experiments.
\subsubsection{Straight Line}
\begin{figure}[!htb]
                \centering
                \includegraphics[width=3.5in]{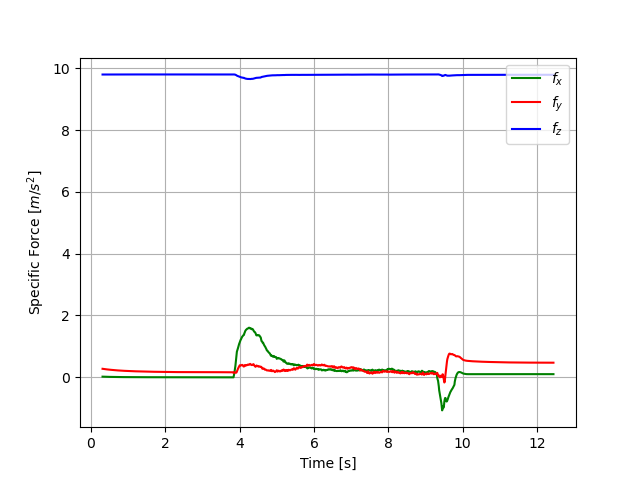}
                \caption{Specific force readings during a straight line recording. The recordings contain three seconds of stationary conditions at the beginning and end of the trajectory.}
                \label{fig_sf_ins}
\end{figure}

\begin{figure}[!htb]
                \centering
                \includegraphics[width=3.5in]{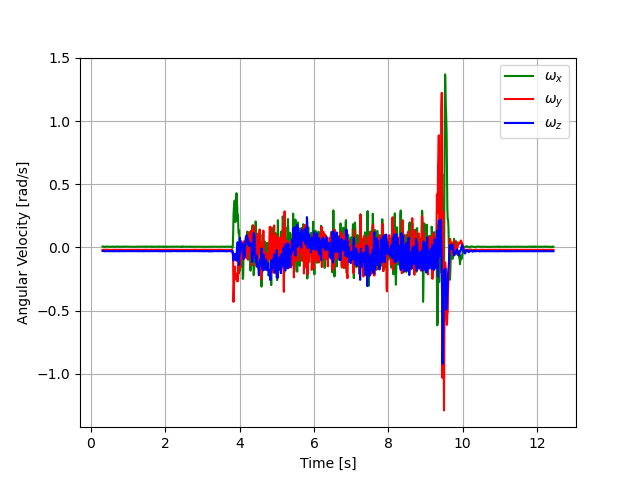}
                \caption{Angular velocity readings during straight line recordings. The recordings contain three seconds of stationary conditions at the beginning and end of the trajectory.}
                \label{fig_gyro_ins}
\end{figure}

To evaluate the INS solution, 24 recordings of driving in a straight line were made with the Samsung Galaxy S8 cellphone as part of the autonomous platform's inertial dataset \cite{dataset2022}. The length of the straight line trajectory was $6.3$m and the recordings were done indoors. 
Each of the recordings contains at least three seconds of stationary conditions at the beginning and end of the trajectory. \\
An example of typical recordings of this trajectory type is presented in Figures \ref{fig_sf_ins} and \ref{fig_gyro_ins} for the accelerometers and gyroscopes, respectively. The direction of the motion is along the $x$-axis, therefore there is a spike in the specific force in that axis at the beginning of the motion, and then decreases towards zero because we tried to keep a constant velocity during the experiments. At the end of the motion, deceleration slowed down the car until it came to a complete halt. A slight force in the $y$ and $x$ axes can be observed as it was difficult to maintain a straight line along the course.
\subsubsection{Periodic Motion - Short Route} \label{subsec:PM-short}
\begin{figure}[!htb]
                \centering
                \includegraphics[width=3.5in]{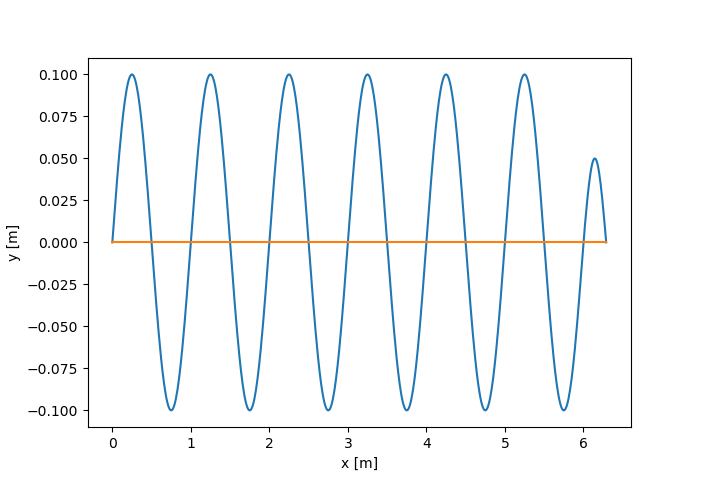}
                \caption{Illustration of the periodic motion and straight line  trajectories.}
                \label{fig_trajectories}
\end{figure}

To evaluate our proposed approach, a sine shaped trajectory was recorded 23 and 30 times with the two smartphones:  Samsung Galaxy s8 and s6, respectively. The start and end points of the trajectory were the same as for the straight line trajectory, with the same distance of $6.3m$. The recordings were done indoors with three seconds of stationary conditions at the beginning and end of the trajectory. An amplitude of approximately $0.1m$ was applied in periods of $1m$ length, with different velocities of the mobile robot. An illustration of this trajectory type with a straight line trajectory is presented in Figure \ref{fig_trajectories}.
In addition, another 26 recordings were gathered using only the Samsung Galaxy S8 smartphone with longer periods of $1.5$m and $0.2$m of amplitude on the same route. \\
The average time of the trajectories with periodic motion is $11$s for $1.5m$ peaks and about $14.5$s for the $1m$ peak trajectories. It is more than twice than in the straight line trajectories, which have an average duration  of $5$s for the same travelled distance (start to end point).
\subsubsection{Periodic Motion - Long Route} \label{subsubsec:LR}
\begin{figure}[!h]
            \centering
            \includegraphics[width=2.5in]{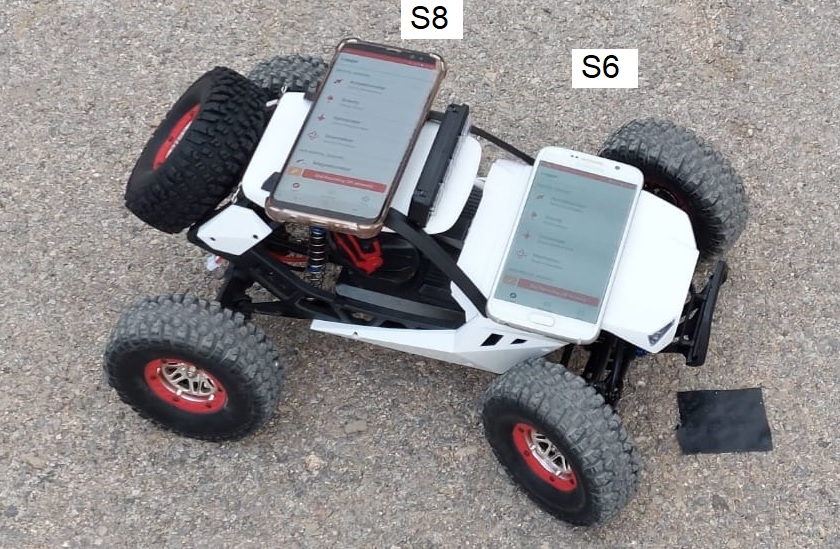}
            \caption{Setup of the RC car with two phones.}
            \label{fig_car_2_phones}
        \end{figure}

\begin{figure}[!htb]
                \centering
                \includegraphics[width=3.5in]{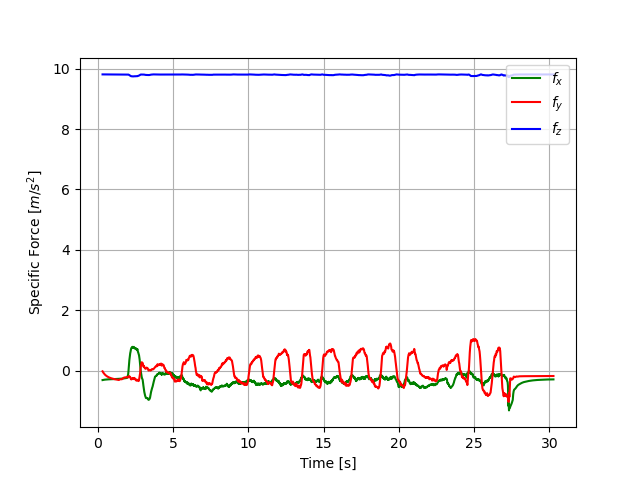}
                \caption{Specific force readings during a periodic motion recording. The recording contains three seconds of stationary conditions at the beginning and end of the trajectory.}
                \label{fig_sf_long}
            \end{figure}

\begin{figure}[!htb]
                \centering
                \includegraphics[width=3.5in]{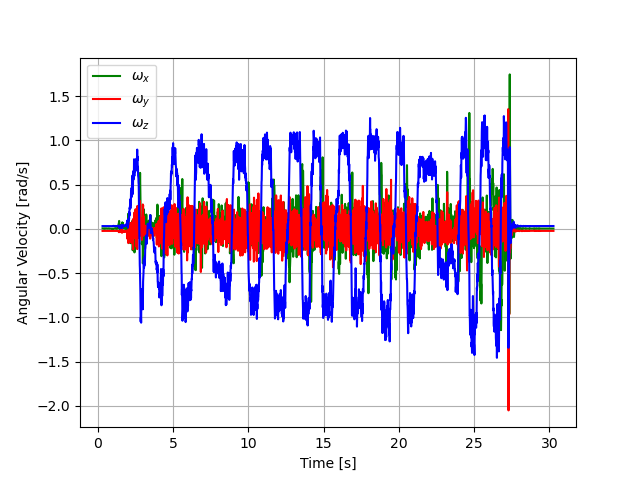}
                \caption{Angular velocity readings during a periodic motion recording. The recording contains three seconds of stationary conditions at the beginning and end of the trajectory.}
                \label{fig_gyro_long}
            \end{figure}

In the same manner, as the short route, a sine shape trajectory was recorded ten times with the Samsung Galaxy s8 and ten times with the Samsung Galaxy s6 for a longer distance of $13$m, which is about twice the short route. An amplitude of approximately $0.1$m was examined with periods of $1 m$, with different velocities. These recordings were taken outdoors. The smartphones were placed together on the car, with the s8 in the same spot as in the short route recordings and the s6 on the front of the car, as shown in Figure~\ref{fig_car_2_phones}. \\
An example of the inertial sensor recordings during this trajectory is presented in Figures \ref{fig_sf_long}-\ref{fig_gyro_long}.  The periodic motion is seen in the specific force $f_y$ and in the gyro $\omega_z$ readings. 

\subsubsection{L-Shaped - Straight Lines}
\hlc{To examine the robustness of our method, an L-shaped trajectory was examined. The trajectory consists  of an $18$m straight line segment followed by a 90 degrees turn and a $10$m straight line segment (L-shape trajectory). This trajectory was carried out on an asphalt surface,  with a slope of approximately $15\degree$ downhill along the first $5$m of the first segment. The total length of the trajectory, $28$m, is more than twice the long route presented in Section} \ref{subsubsec:LR}.
\hlc{Ten recordings were gathered using a Galaxy S6 smartphone. The smartphone was located on the front of the mobile robot similar to the location  used in the long route.}
\subsubsection{L-Shaped - Periodic Motion}
\hlc{The same L-shaped trajectory, as in the previous section, using the Galaxy S6 smartphone, is used. Instead of moving in straight lines, a periodic motion was applied with periods of  $1$m and an amplitude of  $0.1$m. This  trajectory was recorded ten times.}
\subsubsection{Summary}
A \hlc{total of 143 experiments with a total of 75 minutes} were made. Among them, 83 experiments were recorded with the Samsung Galaxy S8 smartphone and \hlc{60} with the Samsung Galaxy S6 smartphone. One hundred and three experiments were made indoors on a floor, while \hlc{40} experiments were recorded outdoors on an asphalt surface. The number of experiments varies between the different types of trajectories because of unusable recordings due to the manual operation of the robot.\\
The dataset is publicly available and can be downloaded from \datasetUrl.\\
The dataset of the periodic movement was split to have a variety of velocities in both train and test sets, where the train was used to determine the gain, and the test to examine our method. The groups were divided almost equally.

\subsection{Indoor Experiments}
\subsubsection{Straight Line Trajectory} \label{subsubsec:SL}
\begin{figure}[!htb]
                \centering
                \includegraphics[width=3.5in]{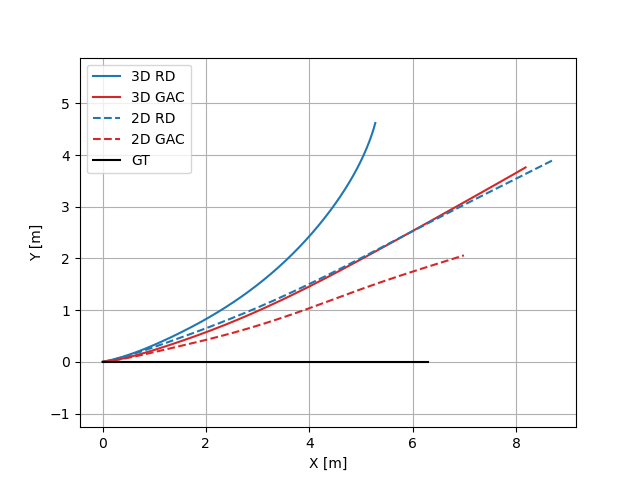}
                \caption{3D and 2D INS trajectories with RD and GAC approaches.}
                \label{fig_ins_routes}
\end{figure}

Equations \eqref{ins_eq1}-\eqref{ins_eq3} were used for calculating the mobile robot location in the INS mechanism. First, the raw inertial sensors readings were plugged into those equations in a naive approach denoted as RD for raw data. Second, to improve performance, zero-order calibration for the gyroscopes was made by utilizing the stationary conditions at the beginning of the trajectory and addressing the mean value in each axis as the bias. In addition, it was assumed that the smartphone is perfectly parallel to the floor, thereby aligning the $z$-axis with the direction of  gravity. As a consequence,  a zero-order calibration was also applied for the accelerometers, taking into account the local gravity value. This gyro and accelerometer calibration approach is denoted as (GAC). The same procedure was applied in the two-dimensional INS mechanism as described in Section~\ref{sec:2dins}.\\
The results with the three and two dimensional INS with the RD and GAC approaches are given in Table~\ref{INS results}. Using the raw data without any calibrations, the 3D INS obtained an error of 3.38m, corresponding to $53.7\%$ of the travelled distance, while the 2D INS obtained a higher error of 3.91m, corresponding to $62\%$. Applying zero order calibration in the GAC approach has less influence over the 3D INS. Yet, the 2D INS error of the travelled distance was reduced from $62\%$ to $28.6\%$. Those results show that after removing the biases of the inertial sensors, the 2D assumptions hold and therefore the performance improves.
\begin{table}[!h]
            \renewcommand{\arraystretch}{1.3}
                \caption{INS errors at the end of the trajectory presented as percentages of the travelled distance.}
                \label{INS results}
                \centering
                \begin{tabular}{|c | c | c|} 
                     \hline
                      & RD & GAC \\
                     \hline
                     3D & 53.7 & 53.1 \\ 
                     \hline
                     2D & 62.0 & 28.6 \\
                     \hline
                \end{tabular}
\end{table}
A typical plot of the 2D and 3D INS solutions with RD and GAC approaches is presented in Figure~\ref{fig_ins_routes} to demonstrate the results discussed above.
\subsubsection{Periodic Motion}
All of the periodic motion recordings were analyzed to extract the peaks, the start point, and the end point of the motion. Then, for each segment, the peak-to-peak distance was calculated for the MoRPI-A method using  \eqref{weinberg} and for MoRPI-G using \eqref{eq:morpi-g}. To calculate the gain of each approach, the training dataset was used with the known travelled distance, allowing to solve for the gain in each equation. The results provided in this section are for the test dataset only. The heading angle at each epoch, $\Delta\psi_k$, was calculated, as in the INS mechanism, using \eqref{ins_eq3}.\\
Besides using the raw data (RD) in MoRPI approaches, we also examined the influence of the gyroscope calibration (GC). Note that when using accelerometer and gyroscopes readings in the INS equations, integration is made on both of the sensor readings, which is the reason that GAC was applied in the straight line trajectories. However, in the proposed MoRPI approaches, the accelerometer readings are used only to detect the peaks and to determine the peak-to-peak distance using an empiric formula without any integration; therefore, only GC was applied.  The calibration process was done in the same manner as the INS method, using the first three seconds of the recordings when the robot was in stationary conditions.\\
Eventually, using the peak-to-peak distance and heading angle, the total distance of the trajectory was calculated using  \eqref{x_dead_reckon}-\eqref{y_dead_reckon}, for both MoRPI approaches.\\
The results of the test dataset of the short route are presented in Table~\ref{weinberg results short} for both smartphones and both MoRPI approaches as a function of the raw data used, RD, or GC, and as the designed peak-to-peak distance. As observed from the table, the proposed MoRPI approaches in all examined configurations greatly improved the 3D and 2D INS solutions. In particular, MoRPI-G obtained the best performance for both smartphone types, with gyro calibration, with a distance error of $4.60\%$-$4.76\%$ compared to the travelled distance. This corresponds to an improvement of the best INS result (2D INS with GAC) by  approximately a factor of five. It is important to note that the variance of the results in each configuration shown in Table~\ref{weinberg results short}  is less than $2.5$cm. \\
The consequence of these results is that the gyroscope is more sensitive than the accelerometer in this process. Thus, peak detection is easier because the peaks are more discernible, so we received more uniform peaks. This also affected the gain calculation in addition to more robustness over different velocities.\\
            \begin{table}[!h]
            \renewcommand{\arraystretch}{1.3}
                \caption{MoRPI errors at the end of the periodic motion trajectory presented as percentages of the travelled distance.}
                \label{weinberg results short}
                \centering
                \begin{tabular}{|c | c | c | c | c | c |} 
                     \hline
                        \multicolumn{2}{| c |}{} & \multicolumn{2}{| c |}{MoRPI-A} & \multicolumn{2}{| c |}{MoRPI-G} \\
                    \hline
                     smartphone & p2p distance & RD & GC & RD & GC\\ 
                    \hline
                     s8 & 1m & 8.25 & 5.87 & 7.94 & 4.76\\ 
                    \hline
                     s8 & 1.5m & 9.84 & 8.25 & 9.05 & 6.67\\ 
                    \hline
                     s6 & 1m & 7.30 & 7.14 & 4.60 & 4.60\\ 
                    \hline
                \end{tabular}
            \end{table}
Table~\ref{tab:comparison methods} summarises the average error in meters for the best approach with each method; i.e., the GAC from 3D and 2D INS, and GC from MoRPI-A and MoRPI-G, relative to the end point of $6.3$m in the $x$-axis, and the improvement in percentages of each method relative to the others. Despite the longer time duration in periodic motion trajectories, compared to the straight line, the position error was significantly lower as described in Section~\ref{subsec:PM-short} and as expected from our analytical assessment in Section~\ref{subsec:AA}.  
            \begin{table}[!h]
            \renewcommand{\arraystretch}{1.3}
                \caption{Comparison between the different methods and our approaches showing the position error.}
                \label{tab:comparison methods}
                \centering
                \begin{tabular}{| c | c | c | c | c |} 
                     \hline
                        Method & Error in meter & 3D INS & 2D INS & MoRPI-A \\
                    \hline
                     3D INS & 3.34 &  &  & \\ 
                    \hline
                     2D INS & 1.80 & 24.5\% &  & \\ 
                    \hline
                     MoRPI-A & 0.37 & 47.2\% & 22.7\% & \\ 
                    \hline
                     MoRPI-G & 0.29 & 48.5\% & 24\% & 1.27\% \\ 
                    \hline
                \end{tabular}
            \end{table}
\subsection{Outdoor Experiments}
\subsubsection{Periodic Motion} To further evaluate our approach, we performed outdoor experiments that differ from the indoor ones, by surface type (asphalt instead of a tiled floor) and  trajectory distance ($13$m instead of $6.3$m). Based on  the analysis of the indoor experiments' results, we examine here only the $1m$ desired peak-to-peak distance using  both smartphones. In addition, the gain that was calculated in the indoor experiments was used for fair comparison; i.e., all outdoor experiments were treated as a new test dataset to examine the robustness of the proposed approach. Finally, we examined the MoRPI-A and MoRPI-G methods with RD and GC approaches, as in the indoor experiments.
\\
The results, presented in Table~\ref{weinberg results long}, show the same behaviour as in the indoor experiments: an improvement when using calibration and an improvement using MoRPI-G, and with similar accuracy.  This is consistent with our assumption of linear error in the proposed method. There is a small difference in the error percentages in the long route, where the main cause is the human factor in the experiments, which becomes more significant as the distance increases. In addition, the poor outdoor conditions, where the ground is not level and included obstacles such as cracks and loose gravel, contributed to the error. Moreover, the setup of the recordings, with two phones recording simultaneously, changed the dynamics of the robot and influenced the results.
        \begin{table}[!ht]
        \renewcommand{\arraystretch}{1.3}
            \caption{MoRPI for long distance errors at the end of the trajectory in percentages.}
            \label{weinberg results long}
            \centering
            \begin{tabular}{|c | c | c | c | c | c |} 
                 \hline
                    \multicolumn{2}{| c |}{} & \multicolumn{2}{| c |}{MoRPI-A} & \multicolumn{2}{| c |}{MoRPI-G} \\
                \hline
                 smartphone & p2p distance & RD & GC & RD & GC\\
                \hline
                 s8 & 1m & 7.15 & 5.62 & 5.85 & 4.46\\ 
                \hline
                 s6 & 1m & 8.62 & 8.46 & 4.77 & 4.15\\ 
                \hline
            \end{tabular}
        \end{table}
\subsubsection{L-Shaped Trajectory}
\hlc{In the same manner as in the straight line trajectory, Section} \ref{subsubsec:SL}, \hlc{the INS equations were used with the two types of configuration: RD or GAC, and 2D or 3D. The error was calculated by the Euclidean distance of the achieved end point relative to the real end point coordinates (28m distance).\\ 
The results, given in Table} \ref{tab:INS_Lshape}, \hlc{show that without any calibration the errors are huge in both cases of pure inertial navigation,  2D and 3D,  with $1523\%$ and $1123\%$, respectively. Using zero-order calibration improves the results to $156\%$ and $161\%$, for 2D and 3D, respectively, but they are still unusable for most applications.}\\
\begin{table}[!h]
            \renewcommand{\arraystretch}{1.3}
                \caption{INS errors at the end of the L-shaped trajectory presented as percentages of the travelled distance.}
                \label{tab:INS_Lshape}
                \centering
                \begin{tabular}{|c | c | c|} 
                     \hline
                      & RD & GAC \\
                     \hline
                     3D & 1123 & 161 \\ 
                     \hline
                     2D & 1523 & 156 \\
                     \hline
                \end{tabular}
\end{table}
\hlc{To evaluate MoRPI approaches on the L-shaped trajectory, the same gains of the long route were used. Thus, all the experiments in this trajectory are addressed as a new test dataset. The performances of the methods MoRPI-A and MoRPI-G with RD and GC were evaluated, as in the preceding experiments, and are presented in Table} \ref{tab:morpi_Lshape}. 
\hlc{The results show the same behaviour as in the indoor and outdoor experiments: an improvement using calibration and improvement using MoRPI-G. In particular, the lowest error, when using a pure inertial navigation approach was 156\% while  MoRPI-G reduced the error to 8.2\%.}\\
\hlc{Focusing on MoRPI approaches, in this experiment, there is a degradation in accuracy compared to the short and long trajectories. The reasons for the degradation are:}
\begin{enumerate}
    \item \hlc{A slope of $15$ degrees in the first $5$m of the trajectory.}
    \item \hlc{The presence of a $90$ degrees  turn.}
    \item \hlc{This experiment included a longer distance than the previous one and the errors caused by the manual operation were more significant.}
\end{enumerate}
\hlc{Despite all of the above issues, and the fact that this experiment was treated as a test, the results show that MoRPI approaches are robust even to complicated scenarios and greatly improve the standalone pure inertial solution.}

\begin{table}[!ht]
        \renewcommand{\arraystretch}{1.3}
            \caption{MoRPI for L-shaped trajectory errors at the end of the trajectory in percentages.}
            \label{tab:morpi_Lshape}
            \centering
            \begin{tabular}{|c | c | c | c | c | c |} 
                 \hline
                    \multicolumn{2}{| c |}{} & \multicolumn{2}{| c |}{MoRPI-A} & \multicolumn{2}{| c |}{MoRPI-G} \\
                \hline
                 smartphone & p2p distance & RD & GC & RD & GC\\
                \hline
                 s6 & 1m & 18.97 & 16.14 & 8.63 & 8.2\\ 
                \hline
            \end{tabular}
\end{table}
\section{Conclusion}\label{sec:con}
To reduce the error drift in situations of pure inertial navigation we proposed MoRPI, a mobile robot pure inertial approach. \hlc{To evaluate MorPI and  baseline} approaches, two different smartphones were mounted on a mobile robot and their inertial sensors were recorded in two different types of periodic motion, differing in surface type and length. A total of \hlc{143 trajectories were recorded with a total time of 75 minutes}. \\ 
Our results showed that the 2D INS with accelerometer and gyroscope calibration obtained the best performance in the baseline approaches, achieving an error of $1.8$m for the $6.3$m trajectory, which corresponds to $28.6\%$ of the travelled distance. Using the MoRPI-A approach, the average error using the two smartphones was $6.5\%$ of the travelled distance, while MoRPI-G obtained the overall best performance with an error of $4.68\%$ of the travelled distance. This means that our proposed approach improved the INS approach by a factor of six.
We showed that even for twice the distance, and as a consequence of a longer duration of movement, the error increased in a linear manner. For example, the error over $6.3$m was $4.76\%$ using the Samsung s8, and over $13$m the error was $4.46\%$.
\hlc{Finally, an L-shaped trajectory, including a slope and a 90 degrees turn, was also examined. As in the other trajectories, MoRPI approaches greatly improved the pure inertial solution.} \\
The above experiment results and characteristics coincide with our analytical assessment closed form solution for the position error of the INS and MoRPI approaches.  
To conclude, in scenarios where pure inertial navigation is needed, our proposed approaches, MoRPI-A and MoRPI-G, provide a lower position error compared to the INS solution. In particular, MoRPI-G obtained the best performance using only the gyroscopes readings. All of the recorded data and code used for our evaluations  are publicly available at \datasetUrl.

\section*{Acknowledgment}
A. Etzion was supported by The Maurice Hatter Foundation. 

\bibliographystyle{IEEEtran}
\bibliography{./bib/references.bib}

\begin{IEEEbiography}[{\includegraphics[width=1in,height=1.25in,clip,keepaspectratio]{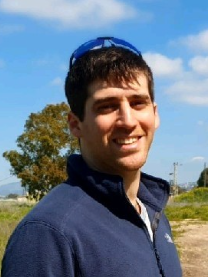}}]{Aviad Etzion}
received a B.Sc. degree in electrical engineering from the Technion-Israel Institute of Technology. He is currently pursuing an M.Sc. degree with the Autonomous Navigation and Sensor Fusion Laboratory, the Hatter Department of Marine Technologies, University of Haifa. His research interests include navigation, deep learning, and sensor fusion.
\end{IEEEbiography}

\begin{IEEEbiography}[{\includegraphics[width=1in,height=1.25in,clip,keepaspectratio]{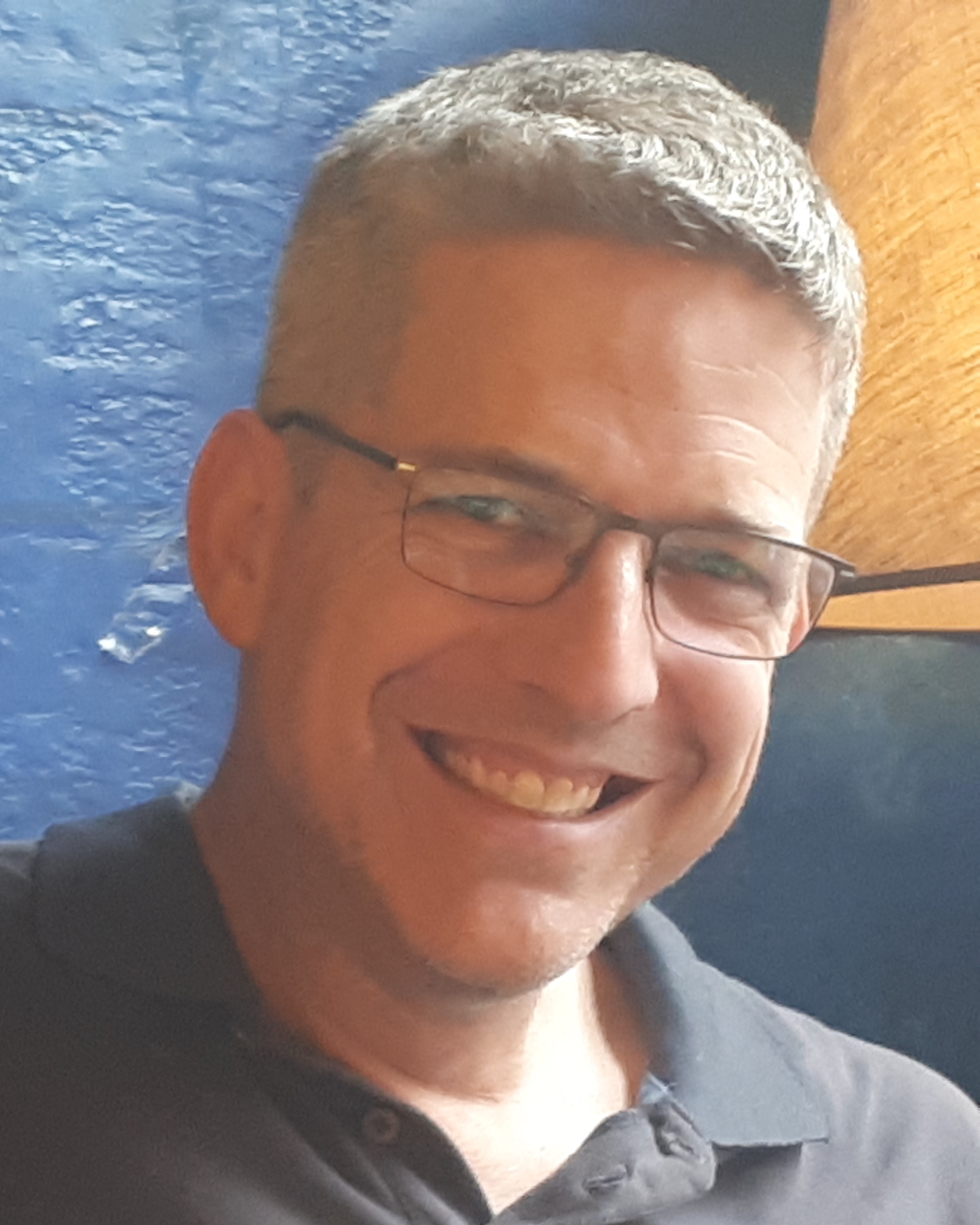}}]{Itzik Klein}
(Senior Member, IEEE) received the B.Sc. and M.Sc. degrees in Aerospace Engineering and a Ph.D. degree in Geo-information Engineering from the Technion - Israel Institute of Technology, Haifa, Israel, in 2004, 2007, and 2011, respectively. He is currently an Assistant Professor, heading the Autonomous Navigation and Sensor Fusion Laboratory, at the Hatter Department of Marine Technologies, University of Haifa, Israel. His research interests include data-driven based navigation, novel inertial navigation architectures, autonomous underwater vehicles, sensor fusion, and estimation theory.
\end{IEEEbiography}

\end{document}